# Data Article template

**Article Title**

LFW-Beautified: A Dataset of Face Images with Beautification and Augmented Reality Filters

**Authors**

Pontus Hedman, Vasilios Skepetzis, Kevin Hernandez-Diaz, Josef Bigun, Fernando Alonso-Fernandez

**Affiliations**

All authors are with the Center for Applied Intelligent Systems Research (CAISR), Halmstad University, SE 301 18 Halmstad, Sweden

**Corresponding author(s)**

Fernando Alonso-Fernandez (feralo@hh.se)

**Abstract**

Selfie images enjoy huge popularity and acceptability in social media platforms. The same platforms centered around sharing this type of images offer different filters to "beautify" them or to incorporate augmented reality effects (like noses or glasses) before they are uploaded. Studies suggests that filtered images are likely to attract more views, comments, and engagement of others [R1]. Selfie images are also in increasing use in security applications due to mobiles becoming data hubs for many transactions [R2]. Also, video conference applications, which have boomed during the pandemic, include beautification or augmented reality filters.

Such filters may destroy or modify biometric features that would allow person recognition or even detection of the face itself, even if such commodity applications are not necessarily used with the purpose of compromising facial systems. This could compromise the normal operation of such systems, or subsequent investigations like crimes in social media [R3], where automatic analysis is usually necessary given the amount of information posted in social sites or stored in devices or cloud repositories.

To help in counteracting the mentioned issues and allow fundamental studies with a common public benchmark, we contribute with a database of facial images that includes the application of several image manipulations like those found in social media. The database includes image enhancement filters (which mostly modify contrast and lightning) as well as augmented reality filters that incorporate items like "Dog nose", "Transparent glasses", "Sunglasses with slight transparency", and "Sunglasses with no transparency" to the face image. Additionally, images with sunglasses are processed with a reconstruction network which has been trained to learn to reverse such modifications. This is because obfuscating the eye region has been observed in the literature to have the highest impact on the accuracy of face detection or recognition [R4].

We start from the popular Labeled Faces in the Wild (LFW) database [R5], to which we apply the different modifications, generating 8 datasets, each one containing one particular modification. Each dataset contains 4,324 images of size 64 x 64, with a total of 34,592 images. The use of a public and widely employed face dataset allows for replication and comparison. We also include the original (unfiltered) images to allow other researchers to incorporate new filters of their choice, or to train different image reconstruction methods.

This article is submitted as a co-submission of another Elsevier journal [R6].

**Specifications Table**

| Subject | Computer Science<br>Computer Vision and Pattern Recognition |
|---|---|
| **Specific subject area** | The dataset can be used for research on face detection or face recognition when typical beautification or augmented reality filters employed in social media are used. It contains visible images of faces. |
| **Type of data** | Image (.png) |
| **How data were acquired** | The data is built from the popular Labeled Faces in the Wild (LFW) database [R5]. Several image manipulations are applied synthetically to simulate beautification or augmented reality (AR) filters found typically in social media applications. For enhancement, we use the 9 most popular selfie Instagram filters [R7], which mostly change contrast and lighting. Regarding AR filters, we simulate "Dog nose", "Transparent glasses", "Sunglasses with slight transparency", and "Sunglasses with no transparency". Additionally, images with sunglasses are processed with a reconstruction network trained specifically to revert such modification. |
| **Data format** | Raw |
| **Parameters for data collection** | The database consists of face photographs of celebrities obtained from the web, containing a large range of variations in pose, lightning, expression, resolution, age, gender, race, occlusions, make-up, accessories, etc. |
| **Description of data collection** | Original images are processed with different modification algorithms. We focus on two manipulations: image enhancement and Augmented Reality (AR). AR filters in particular have not been considered in the face detection/recognition literature previously. |

|  | For enhancement, we use the 9 most popular selfie Instagram filters [R7], which mostly change contrast and lighting. They are recreated using a trained algorithm that learns each modification [R8]. Regarding AR filters, they obfuscate face parts that can be critical for recognition, in particular nose and eyes. We apply: "Dog nose", "Transparent glasses", "Sunglasses-slight transparency", and "Sunglasses-no transparency". These are merged with the face by using the landmarks given by a face detector [R9]. Additionally, images with sunglasses are processed with a modified version of the popular U-NET segmentation network [R10], which has been trained to learn to reverse such modifications. |
|---|---|
| **Data source location** | Institution: School of Information Technology, Halmstad University<br>City/Town/Region: Halmstad<br>Country: Sweden<br>Latitude and longitude (and GPS coordinates, if possible) for collected samples/data: 56° 39' 30.59" N, 12° 52' 25.79" E |
| **Data accessibility** | Repository name: GitHub<br>Data identification number: [provide number, if available]<br>Direct URL to data:<br>https://github.com/HalmstadUniversityBiometrics/LFW-Beautified |
| **Related research article** | Pontus Hedman, Vasilios Skepetzis, Kevin Hernandez-Diaz, Josef Bigun, Fernando Alonso-Fernandez, On the Effect of Selfie Beautification Filters on Face Detection and Recognition, Pattern Recognition Letters. In Press. [R6] |

**Value of the Data**

- The dataset can be used to develop new algorithms for face detection or face recognition when the images are post-processed with beautification or augmented reality filters

- The dataset can be used by scientists in signal/image processing, computer vision, artificial intelligence, pattern recognition, machine learning and deep learning fields

- The provided data can help in developing systems that are resilient to the presence of face beautification or augmented reality filters. Application examples include from normal operation of face detection/recognition systems to crime investigation on social media or automatic pre-filtering of images in personal devices or cloud repositories

- The provided data can be used to recreate new social media filters or image reconstruction methods not considered and benchmark the results against studies made by us or by other users of the database, since we also release the employed images in its original form (unfiltered)

**Data Description**

The database includes face photographs, which are further processed with different modification algorithms. Given an original image, it is processed with 7 different modifications, leading to the 8 different datasets indicated in Table 1 (counting the original images as one dataset). The modification algorithms applied are described in detail in the next section.

| Dataset | Manipulation |
|---|---|
| Benchmark | Original images |
| Dog | Dog nose |
| Glasses | Transparent glasses |
| Instagram | Instagram filters |
| Shades_leak | Shades (95% opacity) |
| Shades_recon_leak | Reconstructed shades (95% opacity) |
| Shades_no_leak | Shades (100% opacity) |
| Shades_recon_no_leak | Reconstructed shades (100% opacity) |

Table 1: Summary description of the 8 datasets generated.

Each dataset contains 4,324 images, with a total of 34,592 images. The images are of size 64 x 64, all in png format. The filenames start with the name of the individual, following by a number, e.g. Adrien_Brody_0001.png, George_W_Bush_0001.png, etc. The database consists of images of celebrities crawled from public websites like news sites. They are well-known to the public and their identity could be easily inferred even if the name did not appear in the filename. Each of the datasets of Table 1 is separated in a different folder, named as indicated in the first column, with the files named exactly equal across folders, so the same image with different modifications can be easily identified.

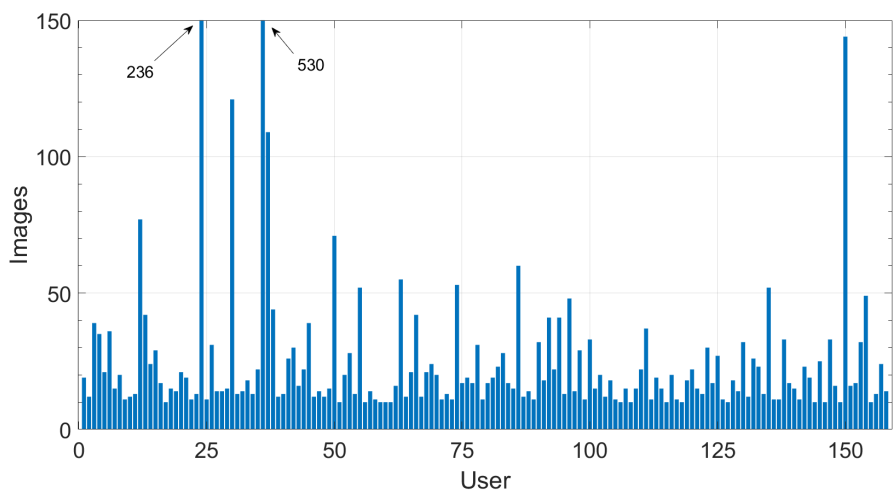

Figure 1: Number of images per individual/per dataset

Figure 1 shows the distribution of images per individual/per dataset (the number of images per dataset is obviously equal, since they are generated by transforming the same image with different modifications). The original database Labeled Faces in the Wild, LFW, [R5] used to generate the mentioned datasets was prefiltered to ensure enough images per person, so only individuals with at least 10 images has been retained. The number of images per individual/per dataset is 27 (average) and 17 (median), with a few persons reaching more than 50, and even several hundreds, as indicated in Figure 1. Examples of images of the different datasets (i.e. with different modifications) are shown in Figure 2.

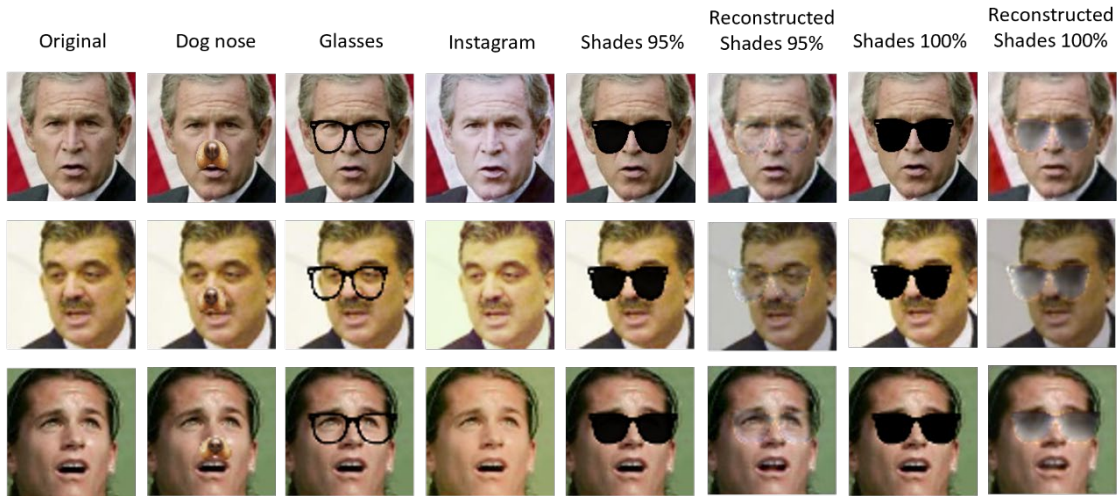

Figure 2: Examples of images from the datasets with different modifications

**Experimental Design, Materials and Methods**

Dataset Composition

Our starting point is the version aligned by funneling [R11] of Labeled Faces in the Wild (LFW) [R5]. The dataset was created at the University of Massachusetts and it is made available for academic purposes. It consists of 13,233 pictures of 5,749 people, with 1,680 individuals having two or more pictures. To ensure a sufficient amount of images per person, we remove people with less than 10 images, resulting in 158 individuals and 4,324 images. Several datasets are then created (Table 1) by processing the 4,324 images available with different modifications.

Application of Image Filters

To our knowledge, there are no datasets available of images with applied beautification filters, which are commonly seen on social media applications. The creation of such a dataset would ideally be made using these applications. However, social media sites such as Instagram or Snapchat do not publicly offer APIs or other means to apply their filters on large amounts of images. Therefore, we have recreated these filters and apply them to our images ourselves.

The filters are applied to the cropped images of the original dataset, obtained by a face detector [R9], which also provides the facial landmarks necessary to incorporate some of the modifications (Figure 3).

If the number of faces found is more than one or if the algorithm does not find a face at all, then the image is discarded.

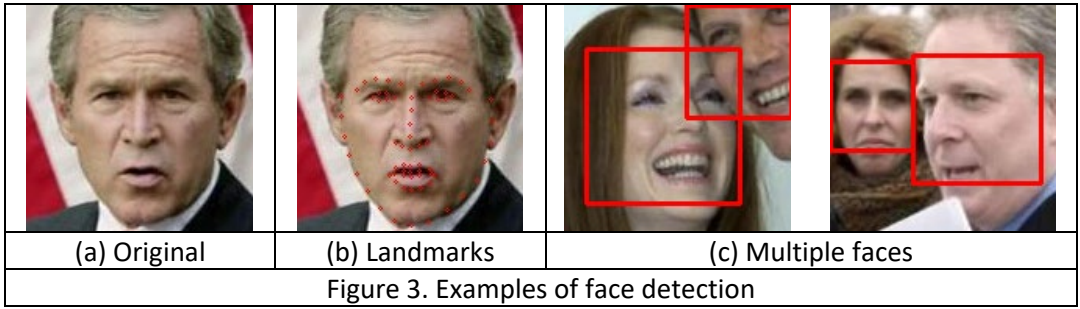

| (a) Original | (b) Landmarks | (c) Multiple faces |

Figure 3. Examples of face detection

*Contrast & Lighting Enhancement filters ("Instagram" filters)*

Instagram is a social media platform centered around the sharing of images. The service offers several ways to filter the images and "beautify" them before upload. A study on another platform, Flickr, found that filtered images are more likely to be viewed and commented, thereby achieving a higher engagement [R1]. The filters offered on Instagram are varied. We focus on the most popular ones, which change the contrast and lighting of the image. This type of filtering is arguable what is most associated with the service today. We choose to use the 9 most "popular" "selfie filters" according to Canvas.com [R7], with the ranking is based on the number of images with a particular filter and the hashtag "#selfie". The applied filters are Aden, Ashby, Dogpatch, Gingham, Hudson, Ludwig, Skyline, Slumber, and Valencia. Figure 4 shows an example of each filter. The Instagram filter images are created using the Instafilter library in Python [R8]. This library uses a four-layer fully connected neural network to learn the RGB, lightness, and saturation changes of various Instagram filters. For each input image of the database, one (1) of the nine (9) most popular "selfie" filters are randomly applied to create the "Instagram" dataset mentioned in Table 1.

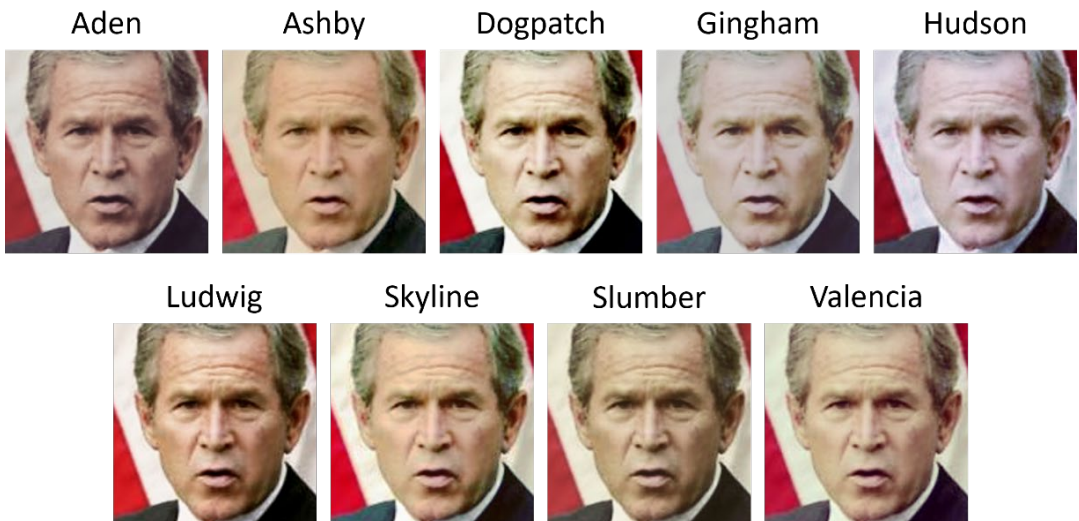

Figure 4: Examples of Contrast & Lighting Enhancement filters ("Instagram" filters)

*Augmented Reality (AR) filters*

We recreate filters observed on various platforms, including social media (like Snapchat) or conference applications (like Zoom). One particularity of some of these filters is that they obfuscate parts of the face, which potentially can make face detection or recognition more difficult. Therefore, we concentrate on filters that hide key areas such as the eyes or the nose. We choose four AR filters to apply to the face images: "Dog nose", "Transparent glasses", "Sunglasses with slight transparency", and "Sunglasses with absolutely no transparency". This results in four additional datasets, as detailed in Table 1. Examples of filtered images are provided in Figure 2. The AR filters are applied based on the calculated landmarks [R9], and the AR add-on is scaled/rotated to size and merged with the target image. Thus, the application of glasses or shades, for example, is done by scaling the original shades image based on the total width of the eyebrows, and the height between the eyebrows and the third quartile of the nose bridge. The filter image is then applied by merging the two images, with the shades centered horizontally with the nose bridge.

Reconstruction of Images with Shades

A reconstruction network has been trained as well to reconstruct the AR shades filters. These two filters have been selected for reconstruction as they completely obfuscate the eye region, which has been observed to have the highest impact on the accuracy of face detection or recognition [R4].

The network used is based on the U-Net network proposed in [R10]. Originally proposed for image segmentation, it outperformed more complex networks in accuracy and speed while requiring less training data. It has a compression or encoding path with convolutions and max-pooling, followed by a decompression or decoding path with up-convolutions. This gives the network a U-shape (Figure 5), hence its name. Another important reason for its success is the residual links that connect maps of the encoding and decoding paths, with channels being concatenated, providing information from various compression stages of the data. In our case, it allows the model to ignore the parts of the image that remain the same and focus on the parts that are modified after applying the filter to the input.

The original network has been modified since our task is different. Inspired by [R12], max-pooling and up-convolutions are changed to strided convolutions/transposed convolutions. Also, map concatenation in residual links is changed by addition to halve the number of channels. With these changes, we expect to still retain changes of image patches while counteracting over-fitting and stabilizing network training. Finally, batch normalization is performed after every second convolution to assist the learning process further. All these mentioned changes are also incorporated to Figure 5.

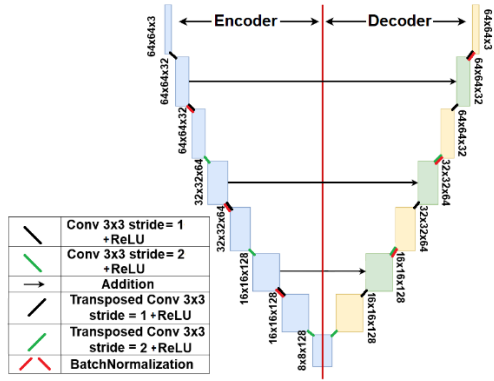

Figure 5: U-NET model employed. The blue rectangles are convolutions (compression), the yellow are transpose convolutions (expansion), and the green symbolizes the addition of blue and yellow through the add operation

To train the reconstruction networks, we use the CelebA database (202,599 pictures of 10,177 people) (Liu et al., 2015), to which we apply the filters *shades_leak* and *shades_no_leak*. Then, the U-NET is trained to revert the modifications and produce the original unmodified image at the output. We use a batch size of 64 and images of 64 x 64 x 3 (similar to the size of our generated datasets), with Adam as optimizer and the MSE between the output and the target (unfiltered) images as loss.

The use of a different database to train U-Net is because we want to avoid that the reconstruction networks 'see' images from the same dataset, allowing to test the generalization ability of the U-Net model on unseen data. After training, the images of our datasets *shades_leak* and *shades_no_leak* are fed to the U-Net in order to be reconstructed. This results in two additional datasets, denoted in Table 1 as *shades_recon_leak* and *shades_recon_no_leak*. Examples of reconstructed images are provided in Figure 2.

Dataset Use

The intended purpose of the dataset is to be used in the training and evaluation face detection or recognition systems when the images are post-processed with beautification or augmented reality filters. Application examples include from normal operation of face detection/recognition systems to crime investigation on social media or automatic pre-filtering of images in personal devices or cloud repositories.

The use of a public and widely employed face dataset allows for replication and comparison, an important aspect of academic research. Releasing the subset of images that we have employed in its original form (unfiltered) would also allow other researchers to incorporate new filters of their choice.

The data contained in the database can be used as-is without filtering or enhancement.

**Ethics Statement**

The data does not include experimentation with human subjects or animals.

Our starting dataset, Labeled Faces in the Wild (LFW), is a public benchmark released in 2007 for studying the problem of unconstrained face recognition. It contains images of famous people well-known to the public, collected over the internet from public websites like news sites. The database is downloadable from the website of the authors via direct download link[1] without any necessary agreement or license. Processed versions of the LFW database like ours have been also made available by other researchers in their own repositories after they carry out subsequent studies, e.g. a new face alignment algorithm (https://talhassner.github.io/home/projects/lfwa/index.html). Such processed versions are even mentioned in the original website of the original LFW database (http://vis-www.cs.umass.edu/lfw/#resources).

**CRediT author statement**

**Pontus Hedman**: Conceptualization, Methodology, Investigation, Data Curation, Writing - Review & Editing

**Vasilios Skepetzis**: Conceptualization, Methodology, Investigation, Data Curation, Writing - Review & Editing

**Kevin Hernandez-Diaz**: Conceptualization, Supervision, Project administration, Writing - Review & Editing

**Josef Bigun**: Conceptualization, Supervision, Project administration, Writing - Review & Editing

**Fernando Alonso-Fernandez**: Conceptualization, Supervision, Funding Acquisition, Writing- Original draft preparation,

**Acknowledgments**

This work has been carried out by P. Hedman and V. Skepetzis in the context of their Master Thesis at Halmstad University (Master's Programme in Network Forensics). Authors K. Hernandez-Diaz, J. Bigun and F. Alonso-Fernandez would like to thank the Swedish Research Council (VR) and the Swedish Innovation Agency (VINNOVA) for funding their research.

**Declaration of Competing Interest**

The authors declare that they have no known competing financial interests or personal relationships which have or could be perceived to have influenced the work reported in this article.

**References**

[R1] Bakhshi, S., Shamma, D., Kennedy, L., Gilbert, E., 2021. Why we filter our photos and how it impacts engagement. Proceedings of the International AAAI Conference on Web and Social Media 9, 12–21

[R2] Rattani, A., Derakhshani, R., Ross, A., 2019. Introduction to Selfie Biometrics. Springer International Publishing, Cham. pp. 1–18

---

[1] The LFW dataset is available at http://vis-www.cs.umass.edu/lfw/

[R3] Powell, A., Haynes, C., 2020. Social Media Data in Digital Forensics Investigations. Springer International Publishing, Cham. pp. 281–303

[R4] Zeng, D., Veldhuis, R., Spreeuwers, L., 2021. A survey of face recognition techniques under occlusion. IET Biometrics 10, 581–606

[R5] Huang, G.B., Ramesh, M., Berg, T., Learned-Miller, E., 2007. Labeled Faces in the Wild: A Database for Studying Face Recognition in Unconstrained Environments. Technical Report 07-49. University of Massachusetts, Amherst

[R6] Hedman, P., Skepetzis, V., Hernandez-Diaz, K., Bigun, J., Alonso-Fernandez, F., On the Effect of Selfie Beautification Filters on Face Detection and Recognition, Pattern Recognition Letters. In Press.

[R7] Canva, 2020. Most popular Instagram filters from around the world. URL: https://www.canva.com/learn/popular-instagram-filters/

[R8] Hoppe, T., 2021. thoppe/instafilter. URL: https://github.com/thoppe/instafilter. original-date: 2020-08-29T13:49:40Z

[R9] Geitgey, A., 2018. face-recognition: Recognize faces from Python or from the command line. URL: https://github.com/ageitgey/face recognition

[R10] Ronneberger, O., Fischer, P., Brox, T., 2015. U-net: Convolutional networks for biomedical image segmentation, in: Navab, N., Hornegger, J., Wells, W.M., Frangi, A.F. (Eds.), Medical Image Computing and Computer-Assisted Intervention – MICCAI 2015, Springer International Publishing, Cham. pp. 234–241

[R11] Huang, G.B., Mattar, M., Lee, H., Learned-Miller, E., 2012. Learning to align from scratch, in: Advances in Neural Information Processing Systems, NIPS

[R12] Springenberg, J.T., Dosovitskiy, A., Brox, T., Riedmiller, M.A., 2015. Striving for simplicity: The all convolutional net, in: Bengio, Y., LeCun, Y. (Eds.), 3rd International Conference on Learning